%% file: root.tex
\documentclass[letterpaper,10pt,conference]{ieeeconf}

\ifdefined\pdfminorversion
\fi

\IEEEoverridecommandlockouts
\usepackage{float}
\usepackage{cuted}
\usepackage{capt-of}
\usepackage{amsmath}
\usepackage{amssymb}
\usepackage{array}
\usepackage{afterpage}
\usepackage{booktabs}
\usepackage{cite}
\usepackage{graphicx}
\usepackage{url}
\usepackage{multirow}
\usepackage{makecell}
\usepackage{algorithm}
\usepackage{algpseudocode}
\usepackage[table]{xcolor}

\DeclareFontFamily{OMS}{cmtt}{}
\DeclareFontShape{OMS}{cmtt}{m}{n}{<->ssub*cmsy/m/n}{}

\newcolumntype{L}[1]{>{\raggedright\arraybackslash}p{#1}}
\newcolumntype{C}[1]{>{\centering\arraybackslash}p{#1}}

\newsavebox{\autoprunetablebox}

\newcommand{\autoprunefittable}[2]{%
  \sbox{\autoprunetablebox}{#2}%
  \ifdim\wd\autoprunetablebox>#1%
    \resizebox{#1}{!}{\usebox{\autoprunetablebox}}%
  \else
    \usebox{\autoprunetablebox}%
  \fi
}

\title{\LARGE \bf
AutoPruner: Agentic Depth Pruning for Vision-Language-Action Models
}

\author{
Tao Lin$^{1,*}$,
Yikeng Chen$^{2,*}$,
Qiwei Liang$^{2,*}$,
Zinan Xie$^{2,*}$,
Zhengtu Liang$^{3}$,
and Renjing Xu$^{2,\dagger}$
\\[3pt]
{\small
$^{1}$Beijing Jiaotong University
\qquad
$^{2}$HKUST (GZ)
\qquad
$^{3}$Shenzhen University
}
\\[-1pt]
{\small
$^{*}$Equal contribution and shared first authorship.
\qquad
$^{\dagger}$Corresponding author.
}
}

\begin{document}

\maketitle

\begin{strip}
    \centering

    \includegraphics[width=\textwidth]{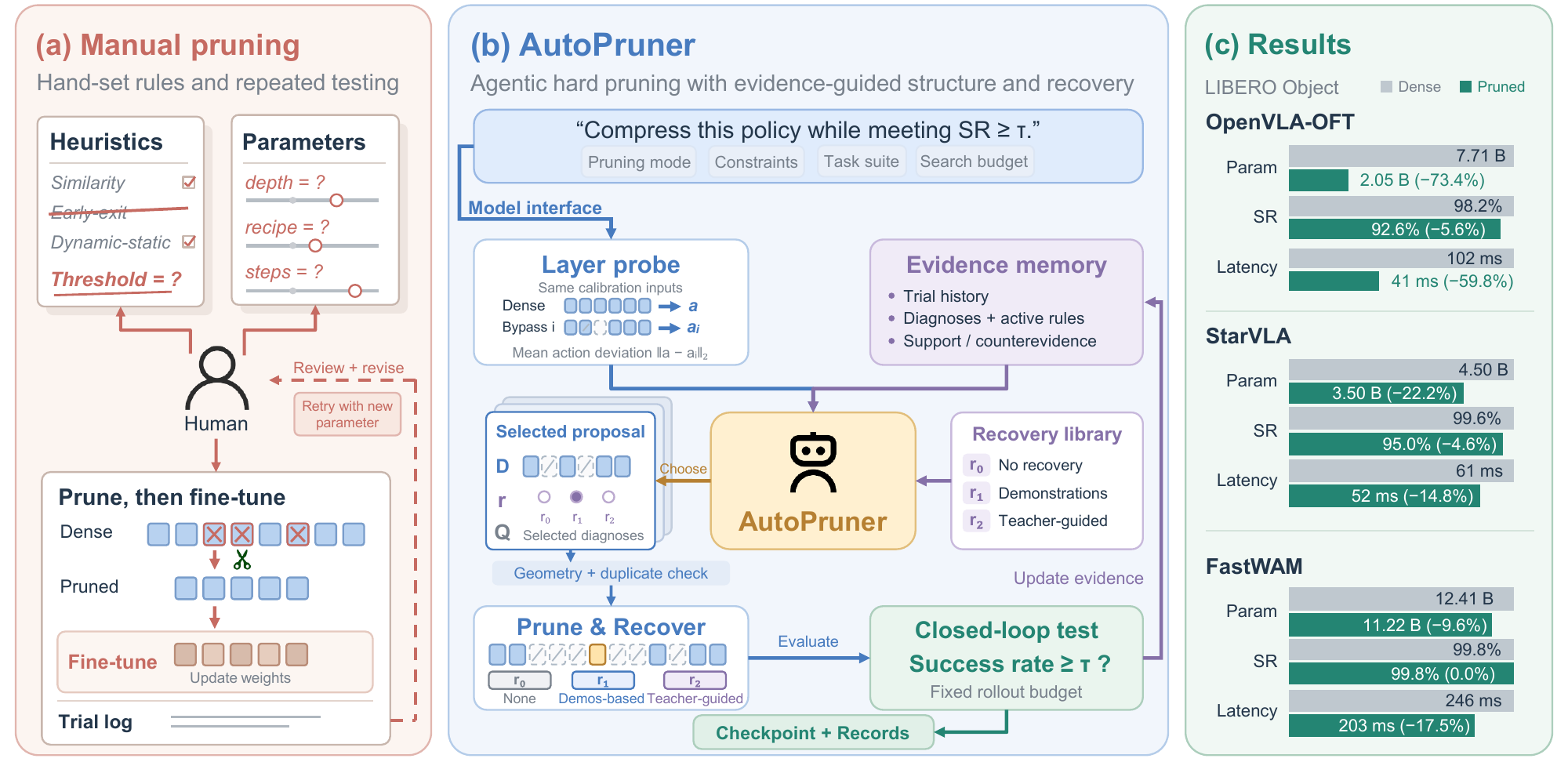}

    \vspace{-3pt}

    \captionof{figure}{
        AutoPruner overview.
        (a) Manual pruning requires repeated human tuning.
        (b) AutoPruner selects pruning structures and recovery strategies
        using probes and closed-loop feedback.
        (c) Results on LIBERO Object show parameter and latency reductions
        across three robot policies.
    }
    \label{fig:teaser}
\end{strip}

\begin{abstract}
\input{sections/abstract}
\end{abstract}

\input{sections/introduction}
\input{sections/related_work}
\input{sections/method}

\input{sections/experiments}

\input{sections/analysis}

\input{sections/discussion}
\input{sections/conclusion}

\raggedbottom

\bibliographystyle{IEEEtran}
\bibliography{references}

\end{document}

%% file: sections/abstract.tex
Layer pruning reduces the size and inference cost of vision-language-action
(VLA) models, but selecting a deployable configuration requires deciding both
which layers to remove and how to recover task performance. These decisions
depend on the model and task, and a failed trial can reflect either the
deletion set or insufficient recovery. We present AutoPruner, an agent-driven
framework that jointly searches pruning structures and recovery recipes
through closed-loop experimentation. Offline layer interventions measure
changes in action output to provide functional evidence. An LLM agent combines
this evidence with evaluation history and retained diagnoses to propose
deletion sets and recovery recipes within a search budget. Each proposal is
executed and evaluated on the target tasks, and the resulting feedback guides
the next experiment. Across OpenVLA-OFT, StarVLA, and FastWAM on four LIBERO
suites, AutoPruner removes 16.7--87.5\% of backbone layers while maintaining
task success above 90\%, with measured inference speedups of up to
2.50$\times$. Further analysis shows that functional probes identify action-sensitive layers despite high representation similarity, and that recovery can make previously failing structures viable.

%% file: sections/introduction.tex
\section{Introduction}

Vision-language-action (VLA) models rely on billion-parameter
backbones whose memory and computation requirements constrain robot
deployment~\cite{kim2024openvla,black2024pi0,starvla2026}. Runtime techniques including token
skipping and computation caching can reduce inference cost without
shrinking the stored model~\cite{jiang2025lightvla,xu2025vlacache}. Layer pruning instead removes transformer
blocks, reducing model size and backbone computation~\cite{men2025shortgpt,yang2025efficientvla,fan2020layerdrop}. The challenge is
finding a deletion set and recovery recipe that preserve task success
on the deployment.

Structure and recovery are coupled. A deletion set that initially disrupts
behavior may become viable after recovery. In our experiments, two deletion
sets exhibit a reversal in observed task performance after the same recovery
procedure. Immediate pruning damage alone therefore cannot determine the
preferred deployment configuration. Feasible structures also vary across
models and tasks: some tolerate long contiguous deletions, while others
require preserving sensitive internal layers. These observations motivate
joint selection of deletion sets and recovery recipes for each deployment.
Single-layer measurements guide this selection, while task evaluations
reveal the combined effect of deleting multiple layers and recovering the policy.

Each trial may require recovery training and hundreds of closed-loop
rollouts. When performance falls short, the next decision is whether to
revise the deletion set or try another recovery recipe. Failure alone cannot
distinguish an unsuitable structure from insufficient recovery; previous
trials provide context for choosing the next step. As
Figure~\ref{fig:teaser}(a) illustrates, manual pruning requires repeated human
intervention to interpret results and rerun experiments. We therefore treat pruning as an iterative experimental process in which previous trials determine the next structure–recovery configuration to evaluate.

We introduce \textbf{AutoPruner}, an agentic framework that jointly searches
deletion sets and recovery recipes for a target deployment, as shown in
Figure~\ref{fig:teaser}(b). Layer interventions measure changes in action
output, providing functional evidence for initial proposals. AutoPruner uses
an LLM agent to integrate functional measurements, trial history, and
diagnoses, jointly revising deletion sets and recovery recipes through
closed-loop task feedback within a search budget. Model-specific tools
execute proposals, and closed-loop evaluation identifies configurations
meeting the target success threshold. The user selects a passing
configuration for export as a physically pruned checkpoint with its
evaluation records.

We evaluate pruning and recovery on OpenVLA-OFT~\cite{kim2025openvlaoft}, StarVLA~\cite{starvla2026}, and FastWAM~\cite{yuan2026fastwam}.
On OpenVLA-OFT Object, the pruned policy retains four of 32 backbone
layers with 92.6\% success, reducing parameters from 7.71B to 2.05B and
latency from 102 to 41 ms. Further analyses examine functional layer evidence
and recovery-dependent pruning depth, while a logged OpenVLA-OFT search
demonstrates proposal revision using task feedback.

Our contributions are threefold:
\begin{itemize}
    \item We formulate deployment-specific VLA depth pruning as a
    sequential decision problem over layer removal and recovery,
    guided by closed-loop task performance under a search budget.

    \item We develop AutoPruner, an agentic framework that integrates
    functional measurements, retained diagnoses, and tool execution
    to revise deletion sets and recovery recipes through experimental feedback.

    \item We evaluate pruning and recovery across three heterogeneous VLA systems
    on four LIBERO suites, removing 16.7--87.5\% of backbone layers
    with inference speedups of 1.11--2.50$\times$ while maintaining
    measured task success above 90\% across all twelve deployments.
\end{itemize}

%% file: sections/related_work.tex
\section{Related Work}

\begin{figure*}[t]
    \centering
    \includegraphics[width=\textwidth]{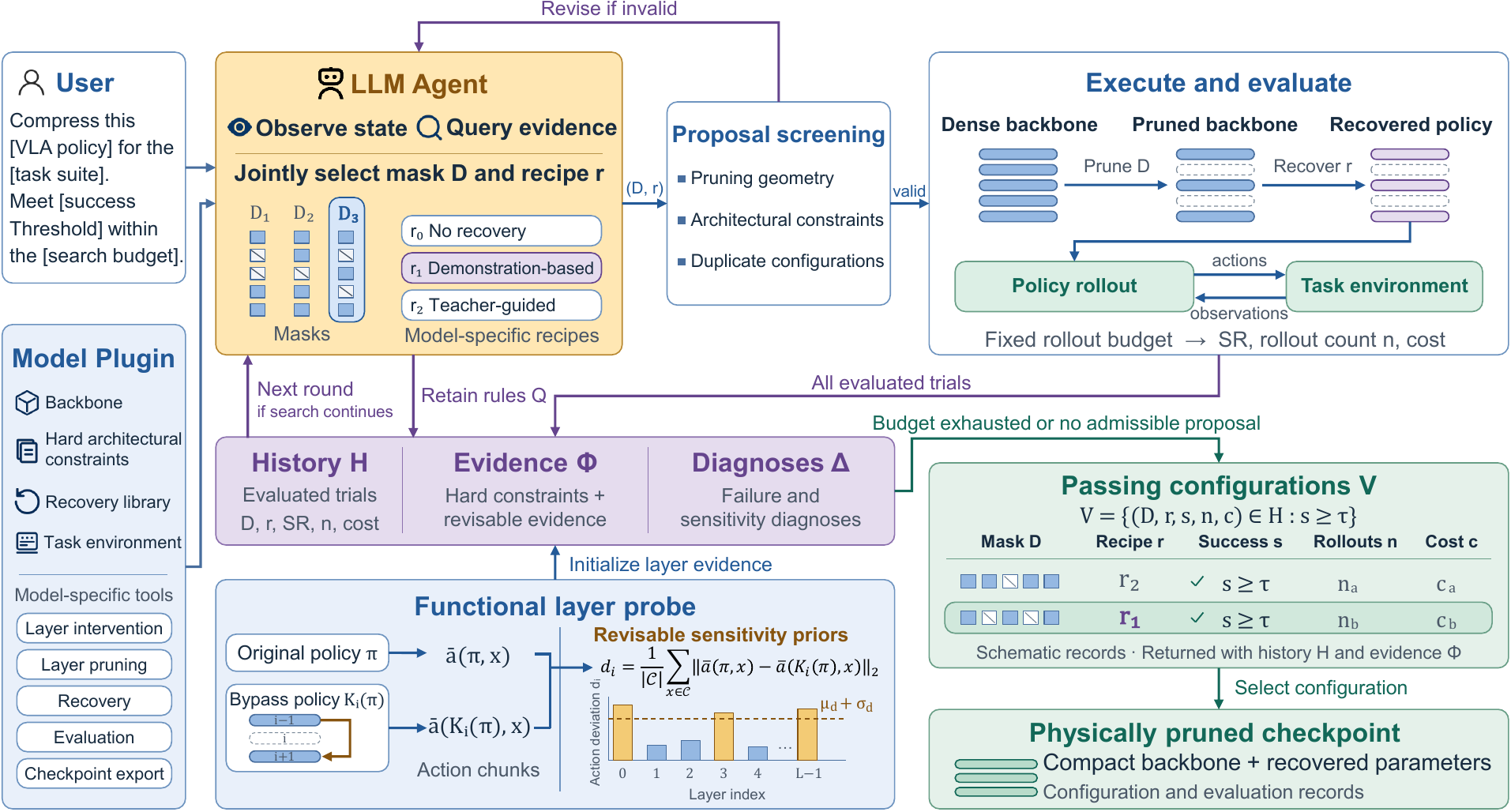}
    \caption{AutoPruner pipeline. Given a target success rate and search budget, an LLM agent uses layer probes and trial history to select pruning masks and recovery strategies. Closed-loop evaluations guide the search, and a selected configuration meeting the target is exported as a physically pruned checkpoint.}
    \label{fig:framework}
\end{figure*}

\subsection{Agentic Model Optimization}
 
ReAct grounds language-model decisions in tool feedback~\cite{yao2023react},
and Reflexion retains textual reflections on task feedback to guide later
attempts~\cite{shinn2023reflexion}. This pattern has been extended to
score-guided optimization, code search, automated experimentation, and full
machine-learning
pipelines~\cite{yang2024opro,jiang2025aide,lu2024aiscientist,ren2026autoresearch,trirat2024automlagent}.
Agentic methods have also been applied to compression. ProfilingAgent reasons
over static cost metrics and runtime profiles to plan layer-wise structured
pruning and post-training quantization~\cite{jafari2025profilingagent}.
Agent-guided pruning selects layers from weight-activation and gradient
sensitivity profiles, using self-reflection and checkpoint rollback across
iterations~\cite{kodathala2026llmcompress}. These systems score a fixed
compression operator against static accuracy or perplexity proxies.
AutoPruner instead jointly selects deletion sets and recovery recipes using
functional measurements and closed-loop task success.
 
\subsection{Efficient Vision-Language-Action Models}
 
Compact VLAs such as TinyVLA and SmolVLA reduce model size or
action-generation costs~\cite{wen2025tinyvla,shukor2025smolvla}, while
OpenVLA-OFT and FAST improve action decoding or
representation~\cite{kim2025openvlaoft,pertsch2025fast}. Quantization reduces
inference memory~\cite{kim2024openvla}, and SQAP-VLA combines it with
visual-token pruning~\cite{fang2025sqapvla}. LightVLA and VLA-Pruner prune
visual tokens~\cite{jiang2025lightvla,liu2025vlapruner}, and VLA-Cache reuses
token features across control steps~\cite{xu2025vlacache}. DeeR-VLA uses
early exits~\cite{yue2024deervla}, while MoLe-VLA and DySL-VLA dynamically
skip layers~\cite{zhang2025molevla,yang2026dyslvla}, retaining backbone layers
in the model. EfficientVLA prunes layers in a training-free manner from input--output
hidden-state similarity and reuses diffusion-head
features~\cite{yang2025efficientvla}.
 
Removed capacity is typically restored by retraining a small set of
parameters or by distilling from the dense
model~\cite{houlsby2019adapters,hinton2015distillation,rusu2016policydistillation},
as in LLM-Pruner, which pairs structural pruning with low-rank
adaptation~\cite{ma2023llmpruner}. GLUESTICK recovers pruned VLA policies
through training-free weight-space correction~\cite{jabbour2025dontrunscissorspruning}.
These methods apply recovery after the deletion set is fixed. AutoPruner
incorporates recovery into the search, evaluating deletion sets together with
recovery recipes on the target tasks.
 
AutoPruner proposes deletions from the action changes caused by individual
layer bypasses, and exports policies meeting the deployment success threshold
with the selected layers physically removed.

%% file: sections/method.tex
\section{Method}

As shown in Figure~\ref{fig:framework}, AutoPruner closes the pruning
and recovery loop with an LLM agent. The LLM agent acts as the proposal controller, selecting the next deletion set and recovery recipe from the accumulated evidence and recovery library. Tools prune, recover, and
evaluate each policy. Feedback guides subsequent proposals toward deeper
pruning under a success threshold and a round budget.

\subsection{Problem Formulation and Agent State}
\label{sec:method_overview}

For a VLA policy $\pi$ with $L$ backbone layers, a deletion set
$D\subseteq\{0,\ldots,L-1\}$ produces the pruned policy $\pi_D$.
A recovery recipe $r$ from the model-specific library $\mathcal{L}(\pi)$
then yields $\pi_{D,r}=R_r(\pi_D)$.
Given a target task suite $\mathcal{S}$ and success threshold $\tau$,
we maximize the number of removed layers subject to this threshold:
\begin{equation}
\max_{D,r}\ |D|
\quad \mathrm{s.t.}\quad
\mathrm{SR}(\pi_{D,r},\mathcal{S})\geq\tau,\qquad
r\in\mathcal{L}(\pi).
\label{eq:objective}
\end{equation}
Here, $\mathrm{SR}$ denotes closed-loop task success rate.
The user allows at most $T$ rounds, each covering proposal generation,
revision, and evaluation of valid proposals.
Rounds without valid proposals also count toward this limit.

Each model plugin specifies the backbone, architectural constraints,
recovery library, and evaluation environment, and provides intervention,
pruning, recovery, evaluation, and export tools.
The agent chooses the deletion set and recovery recipe, the tools execute
them, and a fixed protocol determines acceptance. At round $t+1$, the agent receives
\begin{equation}
    \sigma_t=(H_t,\Phi_t,\Delta_t),
    \label{eq:search_state}
\end{equation}
where the history $H_t$ records evaluated deletion sets, recovery recipes,
success rates, rollout counts, and search costs, and $\Delta_t$ contains
the latest diagnoses.
The evidence store $\Phi_t$ separates mandatory constraints
$\Phi_t^{\mathrm{hard}}$, which define valid proposals, from revisable
evidence $\Phi_t^{\mathrm{ev}}$, which guides selection.
The latter includes layer sensitivities, tested boundaries, and recovery
outcomes. Invalid proposals and rejection reasons are logged separately,
without evaluation or an assigned success rate.

\subsection{Layer Sensitivity Estimation}
\label{sec:evidence}

Layer probes provide initial evidence for agent layer selection.
For each backbone layer $i$, intervention $K_i$ replaces
$h_{i+1}=F_i(h_i)$ with $h_{i+1}=h_i$, where $h_i$ is the layer input.
Each intervention is tested independently with all other parameters fixed.
Using the same model-specific calibration set $\mathcal{C}$, we compare
flattened action chunks from the original and intervened policies
$\bar a(\pi,x)$ for each input $x$:
\begin{equation}
d_i =
\frac{1}{|\mathcal{C}|}
\sum_{x\in\mathcal{C}}
\left\|
\bar a(\pi,x)-\bar a(K_i(\pi),x)
\right\|_2 .
\label{eq:knockout_score}
\end{equation}
These scores rank layers within each model by the action changes caused by removal.
Computing them requires one pass over the calibration set for the
dense policy and one per intervened layer, without training or environment rollouts.

Layers exceeding the mean sensitivity by more than one standard deviation
provide a retention prior. The agent can test their removal and revise
this prior using closed-loop results. Before search, we inspect the profile to select and record a pruning
mode in the model plugin. A low-sensitivity interior bounded by sensitive
layers motivates contiguous deletion, whereas scattered sensitive layers
motivate noncontiguous deletion. The agent chooses deletion sets within
this mode using sensitivities and task feedback.

For the set $\mathcal{C}_k$ of admissible $k$-layer deletion sets,
the reference proposal is
\begin{equation}
D_k^{\mathrm{ref}}
=
\underset{D\in\mathcal{C}_k}{\arg\min}
\sum_{i\in D} d_i .
\label{eq:contiguous_candidate}
\end{equation}
Each admissible set satisfies the selected mode and architectural constraints.
The summed sensitivity provides an initial candidate over contiguous windows
or admissible noncontiguous subsets. Closed-loop evaluation establishes
candidate feasibility and guides subsequent agent revisions.

\subsection{Coupled Structure and Recovery}
\label{sec:action}

\begin{figure}[t]
\centering
\includegraphics[width=\columnwidth]{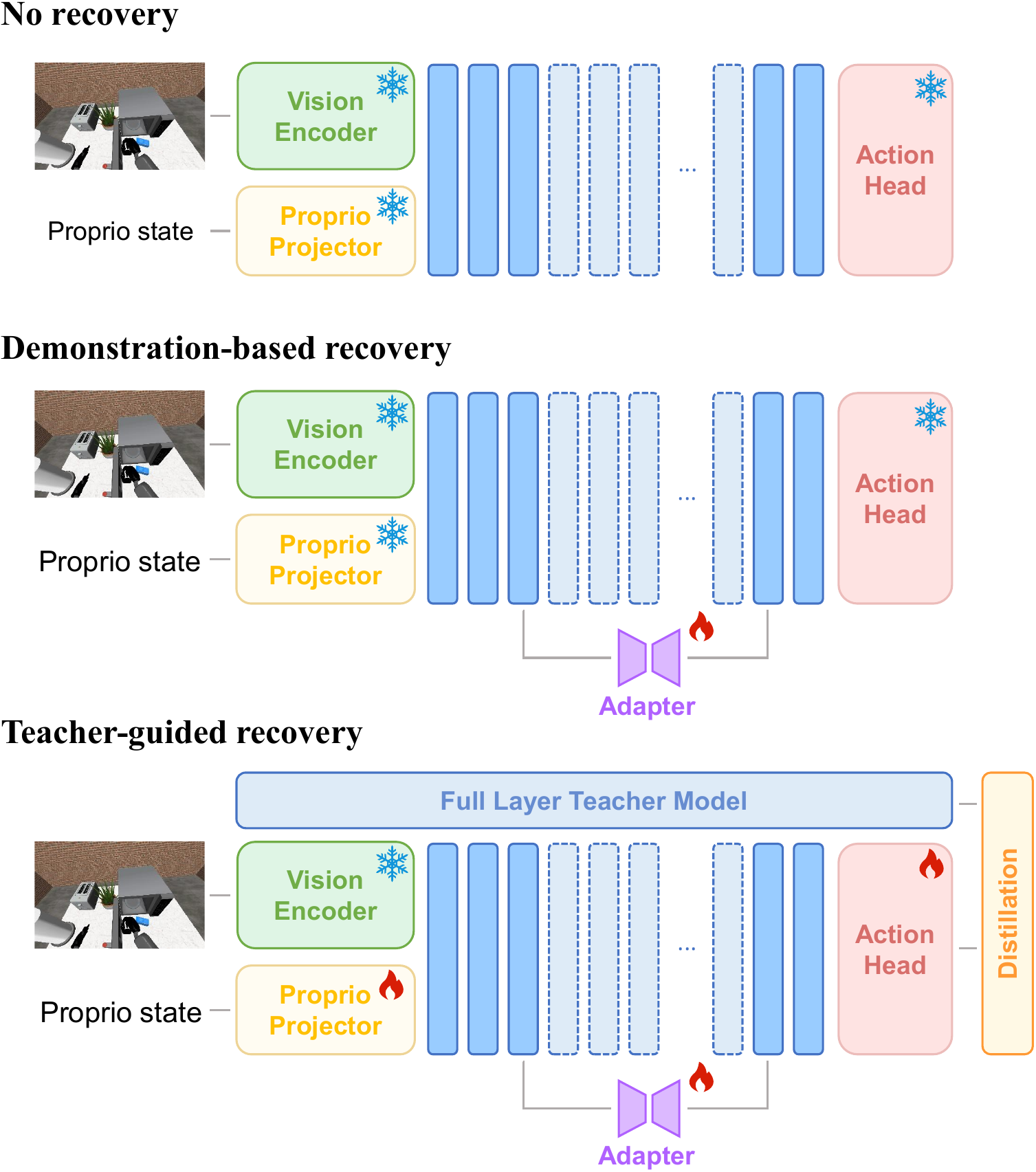}
\caption{OpenVLA-OFT recovery recipes. The three levels evaluate the
pruned policy directly ($r_0$), adapt it to demonstrations ($r_1$), or
add supervision from the dense teacher ($r_2$).}
\label{fig:recovery_recipes}
\end{figure}

The agent uses prior outcomes to jointly select a deletion set and recovery
recipe from three categories, each supporting multiple recipes.
No recovery ($r_0$) evaluates the pruned policy. Demonstration-based recovery
($r_1$) trains on offline demonstrations with the model-specific
action-prediction objective. Teacher-guided recovery ($r_2$) adapts it with
dense-policy supervision~\cite{hinton2015distillation,rusu2016policydistillation}.
Each recipe specifies trainable components and supervision targets.
Figure~\ref{fig:recovery_recipes} shows the OpenVLA-OFT recipes.
Model-specific recipes and training settings appear in Experiments.

Each pair $(D,r)$ is a candidate; recipes can be selected in any order.
Trial feedback guides two complementary moves: revise the deletion
set or adapt the same structure with another recipe. This couples pruning
depth with the recovery needed to regain task performance.

\subsection{Agent-Guided Search and Evaluation}
\label{sec:transition}

At each round, the LLM agent receives the state, pruning mode, and recovery
library as text.
Before proposing, it may query precomputed metrics for selected layers
from the catalog, adding results to its context.
It returns a deletion set, recovery recipe, selected diagnoses, and
experimental rationale in JSON.
The program checks geometry and duplicate $(D,r)$ pairs, returning
rejection reasons for revision. Each round allows up to three measurement-query
rounds and three validation attempts. Without a valid proposal within
these limits, the program advances to the next round.
An unsupported recovery identifier defaults to the first available recipe,
and the executed choice is recorded.

Each valid configuration is evaluated on the target suite under a fixed
rollout budget and passes when $\widehat{\mathrm{SR}}\geq\tau$.
95\% Wilson intervals~\cite{wilson1927interval} summarize sampling uncertainty.

After evaluation, predefined diagnostic rules use the result and history
to identify severe failure without recovery, insufficient recovery, and
sensitivity to deletion changes. For example, with recovery fixed, if
extending $D$ to $D\cup\{i\}$, where $i\notin D$, reduces success by at least
10 percentage points, layer $i$ is flagged as potentially sensitive.

In the next round, the agent selects which diagnoses to retain as rules
in $\Phi^{\mathrm{ev}}$.
Each retained rule stores a problem, suggested action, evidence, scope,
and a heuristic reliability score.
For a layer-retention rule, later trials removing that layer add support
when success falls more than 10 percentage points below the dense
baseline, and counterevidence otherwise.
This boundary differs from the acceptance threshold $\tau$.
For layer-retention rules, Laplace-smoothed outcome counts~\cite{poole2023artificialintelligence}
update reliability, allowing counterexamples to lower scores and retire rules.
The store uses these scores to resolve conflicts and select active rules.
The agent uses these rules and trial history to revise the next pruning
and recovery proposal, closing the experimental loop.

Algorithm~\ref{alg:autoprune} summarizes the search, with admissible
proposals $\mathcal{A}$ and a subset $Q$ of the latest diagnoses selected
by the agent.
The agent step includes queries and validation. Here, $s$, $n$, and $c$
denote success rate, rollout count, and cost.
\begin{algorithm}[t]
\caption{AutoPruner search}
\label{alg:autoprune}
\begin{algorithmic}[1]
\Require Policy $\pi$, suite $\mathcal{S}$, threshold $\tau$, round limit $T$
\Require Recovery library $\mathcal{L}(\pi)$, evidence, mode, constraints
\State $H,\Delta\gets\varnothing$; initialize $\Phi$ from inputs
\For{$t=1,\ldots,T$}
    \State $\mathcal{A}\gets\operatorname{Available}(\Phi,\mathcal{L}(\pi),H)$
    \State \textbf{if} $\mathcal{A}=\varnothing$ \textbf{then break}
    \State $(D,r,Q)\gets\operatorname{Agent}(H,\Phi,\Delta,\mathcal{A})$
    \If{no valid proposal after bounded retries}
        \State Log proposals and rejections; \textbf{continue}
    \EndIf
    \State $\pi_{D,r}\gets R_r(\operatorname{Prune}(\pi,D))$
    \State $(s,n)\gets\operatorname{Evaluate}(\pi_{D,r},\mathcal{S})$
    \State Measure cost $c$; append $(D,r,s,n,c)$ to $H$
    \State $\Phi\gets\operatorname{UpdateRules}(\Phi,Q,H)$
    \State $\Delta\gets\operatorname{Diagnose}(H)$; persist $H,\Phi,\Delta$
\EndFor
\State \Return $\{(D,r,s,n,c)\in H:s\geq\tau\},H,\Phi$
\end{algorithmic}
\end{algorithm}

Search ends after $T$ rounds or when no admissible proposal remains.
Completed trials are saved to preserve results if search is interrupted.
The user receives passing configurations with their compression,
performance, and cost records and selects one for deployment.
AutoPruner exports a physically shortened backbone with recovered parameters
and the selected configuration and evaluation records.

%% file: sections/experiments.tex
\begin{table*}[t]
\centering
\definecolor{aploss}{RGB}{145,0,0}
\definecolor{apkeep}{RGB}{30,125,85}
\providecommand{\sr}[2]{#1\,{\textcolor{aploss}{\scriptsize($-$#2)}}}
\caption{Pruning performance and deployment efficiency on LIBERO~\cite{liu2023libero}.
Arrows show values before and after pruning.
Parentheses show SR changes in percentage points.}
\label{tab:compression_results}
\label{tab:deployment_efficiency}
\footnotesize
\setlength{\tabcolsep}{1pt}
\renewcommand{\arraystretch}{1.22}
\begin{tabular*}{\textwidth}{@{\extracolsep{\fill}}llcccccccc@{}}
\toprule
\multicolumn{4}{c}{} &
\multicolumn{2}{c}{Task performance} &
\multicolumn{4}{c}{Deployment efficiency} \\
\cmidrule(lr){5-6}\cmidrule(l){7-10}
Model & Suite & Layers & Recovery & Dense SR (\%) & Pruned SR (\%) & Pruned GFLOPs & Params (B) & Latency (ms) & Speedup \\
\midrule
\multirow{4}{*}{OpenVLA-OFT~\cite{kim2025openvlaoft}}
 & Object & $32\rightarrow\textcolor{apkeep}{4}$ & Teacher-guided ($r_2$) & 98.2 & \sr{92.6}{5.6} & 1968 & 7.71\,$\rightarrow$\,2.05 & 102\,$\rightarrow$\,41 & 2.50 \\
 & Spatial & $32\rightarrow\textcolor{apkeep}{24}$ & Teacher-guided ($r_2$) & 94.0 & \sr{92.8}{1.2} & 6860 & 7.71\,$\rightarrow$\,6.10 & 97\,$\rightarrow$\,81 & 1.20 \\
 & Goal & $32\rightarrow\textcolor{apkeep}{24}$ & Teacher-guided ($r_2$) & 98.2 & \sr{94.2}{4.0} & 6740 & 7.71\,$\rightarrow$\,6.10 & 96\,$\rightarrow$\,80 & 1.20 \\
 & Long & $32\rightarrow\textcolor{apkeep}{24}$ & Teacher-guided ($r_2$) & 95.2 & \sr{91.2}{4.0} & 6800 & 7.71\,$\rightarrow$\,6.10 & 101\,$\rightarrow$\,83 & 1.21 \\
\midrule
\multirow{4}{*}{StarVLA~\cite{starvla2026}}
 & Object & $36\rightarrow\textcolor{apkeep}{26}$ & Teacher-guided ($r_2$) & 99.6 & \sr{95.0}{4.6} & 1610 & 4.50\,$\rightarrow$\,3.50 & 61\,$\rightarrow$\,52 & 1.17 \\
 & Spatial & $36\rightarrow\textcolor{apkeep}{26}$ & Teacher-guided ($r_2$) & 93.2 & \sr{90.2}{3.0} & 1610 & 4.50\,$\rightarrow$\,3.50 & 60\,$\rightarrow$\,51 & 1.17 \\
 & Goal & $36\rightarrow\textcolor{apkeep}{26}$ & Teacher-guided ($r_2$) & 98.2 & \sr{92.0}{6.2} & 1610 & 4.50\,$\rightarrow$\,3.50 & 60\,$\rightarrow$\,51 & 1.17 \\
 & Long & $36\rightarrow\textcolor{apkeep}{28}$ & Demonstration-based ($r_1$) & 94.8 & \sr{93.8}{1.0} & 1696 & 4.50\,$\rightarrow$\,3.70 & 59\,$\rightarrow$\,53 & 1.11 \\
\midrule
\multirow{4}{*}{FastWAM~\cite{yuan2026fastwam}}
 & Object & $30\rightarrow\textcolor{apkeep}{24}$ & Demonstration-based ($r_1$) & 99.8 & 99.8\,{\scriptsize($-$0.0)} & 2512 & 12.41\,$\rightarrow$\,11.22 & 246\,$\rightarrow$\,203 & 1.21 \\
 & Spatial & $30\rightarrow\textcolor{apkeep}{24}$ & Demonstration-based ($r_1$) & 98.0 & \sr{94.4}{3.6} & 2512 & 12.41\,$\rightarrow$\,11.22 & 246\,$\rightarrow$\,203 & 1.21 \\
 & Goal & $30\rightarrow\textcolor{apkeep}{24}$ & Demonstration-based ($r_1$) & 96.0 & \sr{93.7}{2.3} & 2512 & 12.41\,$\rightarrow$\,11.22 & 246\,$\rightarrow$\,203 & 1.21 \\
 & Long & $30\rightarrow\textcolor{apkeep}{25}$ & Demonstration-based ($r_1$) & 94.6 & \sr{91.6}{3.0} & 2566 & 12.41\,$\rightarrow$\,11.42 & 247\,$\rightarrow$\,211 & 1.17 \\
\bottomrule
\end{tabular*}
\end{table*}
\section{Experiments}

\subsection{Experimental Setup}

We evaluate pruning and recovery on OpenVLA-OFT~\cite{kim2025openvlaoft}, StarVLA~\cite{starvla2026}, and FastWAM~\cite{yuan2026fastwam} across the
Object, Spatial, Goal, and Long suites of LIBERO~\cite{liu2023libero}. Each model--suite pair is a deployment with its dense policy as the reference.

Recorded OpenVLA-OFT agent runs span 4--6 proposal rounds, with GPT-5.5
using evaluation feedback to revise pruning and recovery configurations.
The Object case study evaluates four proposals over 200 episodes each (800 total).
Search ends at the round limit or when no admissible proposal remains,
returning all passing configurations with performance and cost records
for deployment selection.

We report empirical closed-loop success rate (SR) over 500 episodes per main
evaluation. OpenVLA-OFT uses the first 50 provided initial states per task
for final evaluation. Search and final evaluation use disjoint initial
states drawn from the same task initialization distribution.
We report layer counts, parameters, GFLOPs, latency, and peak GPU
memory. Latency averages 50 requests after 10 warmup requests;
speedup divides dense by pruned mean latency.

Functional profiling requests 24 calibration inputs per model from LIBERO Long for
OpenVLA-OFT and StarVLA and LIBERO Object for FastWAM. Each model reuses
them for dense and single-layer predictions.
Reusing inputs isolates prediction changes induced by each layer intervention.

Table~\ref{tab:compression_results} groups recovery into
no recovery ($r_0$), demonstration-based recovery ($r_1$), and
teacher-guided recovery ($r_2$). Teacher-guided recovery adds dense-policy
supervision and may also use demonstrations. StarVLA Object uses a separate
ten-layer-pruned checkpoint recovered on Long, without Object-specific recovery training.
OpenVLA-OFT jointly trains the boundary adapter, action head, and
proprioceptive projector with demonstration action L1 loss and teacher
action-token KL divergence. StarVLA trains boundary adapters with
demonstration action L1 loss; its joint recipe also updates the retained
backbone and action head, adding teacher action L1 loss and hidden-state
MSE at corresponding action-token readout positions.
FastWAM fine-tunes the retained video/action MoT and proprioceptive encoder
with video and action flow-matching losses~\cite{lipman2023flowmatching,yuan2026fastwam} on demonstrations.
For the selected Long configurations, StarVLA trains boundary adapters
for 10000 steps with the backbone and action head frozen, while FastWAM
fine-tunes the full MoT for 150 steps. Both use training seed 42,
with effective batch sizes of 8 and 64, respectively.

\subsection{Pruning Performance and Deployment Efficiency}

As shown in Table~\ref{tab:compression_results}, the evaluated configurations remove 16.7--87.5\% of backbone layers while maintaining
over 90\% success across all twelve deployments. OpenVLA-OFT Object achieves the largest
deployment gains: retaining only four of 32 layers with 92.6\% success
reduces parameters from 7.71B to 2.05B and mean latency from 102 to 41 ms,
a 2.50$\times$ speedup computed from unrounded means.
The other OFT suites retain 24 layers. StarVLA and FastWAM also reduce latency
with shallower pruning. Both prune less deeply on Long than on other suites,
with 93.8\% success for StarVLA and 91.6\% for FastWAM.
The retained depth also varies across suites for the same backbone, confirming that feasible pruning is task-dependent.

\begin{table*}[t]
\centering
\caption{VLA acceleration on OpenVLA-OFT Object.
Latency is averaged over 50 inference requests after 10 warmup requests.}
\label{tab:efficiency_comparison}
\footnotesize
\setlength{\tabcolsep}{4pt}
\renewcommand{\arraystretch}{1.15}
\begin{tabular*}{\textwidth}{@{\extracolsep{\fill}}lcccccc@{}}
\toprule
Method & SR (\%) & Latency (ms) & Speedup & GFLOPs & Memory (GiB) & Physical pruning \\
\midrule
OpenVLA-OFT~\cite{kim2025openvlaoft} & 98.2 & 102 & 1.00 & 8876 & 16.63 & No \\
LightVLA~\cite{jiang2025lightvla} & 99.0 & 58 & 1.76 & 2997 & 14.50 & No \\
DySL-VLA~\cite{yang2026dyslvla} & 98.2 & 58 & 1.75 & 4196 & 15.04 & No \\
FastV~\cite{chen2024fastv} & 78.6 & 70 & 1.45 & 4281 & 14.59 & No \\
VLA-Pruner~\cite{liu2025vlapruner} & 98.6 & 72 & 1.41 & 4281 & 14.85 & No \\
VLA-Cache~\cite{xu2025vlacache} & 98.4 & 101 & 1.01 & 6961 & 15.95 & No \\
\midrule
\textbf{AutoPruner} & \textbf{92.6} & \textbf{41} & \textbf{2.50} & \textbf{1968} & \textbf{4.10} & \textbf{Yes} \\
\bottomrule
\end{tabular*}
\par\smallskip
\begin{minipage}{\textwidth}
\footnotesize
Profiling repeats a fixed LIBERO observation on RTX 4090 GPUs, one process per GPU.
VLA-Cache retains its cache between requests. Speedups use the measured dense mean;
memory is peak PyTorch allocation. Dense OpenVLA-OFT SR follows
Table~\ref{tab:compression_results}; other SRs retain their original evaluation sources.
\end{minipage}
\end{table*}
OpenVLA-OFT and StarVLA profiles use recovered checkpoints.
OFT Object uses the same latency, GFLOPs, and dense profiling reference as
Table~\ref{tab:efficiency_comparison}. StarVLA GFLOPs are structural
estimates; its latency and parameter counts include recovery modules.
FastWAM Object, Spatial, and Goal share a six-layer-deletion recovery
checkpoint and fixed-shape profile. Its Long profile reconstructs the five-layer deletion set on the publicly released weights.

\begin{figure*}[t]
\centering
\includegraphics[width=\textwidth]{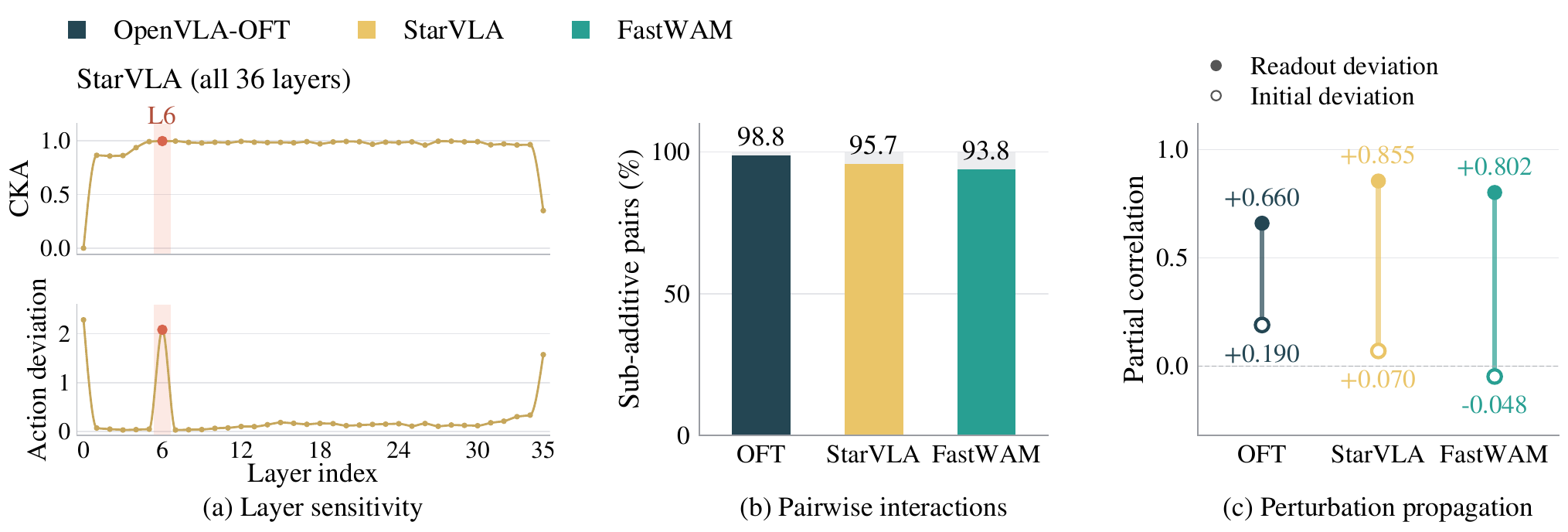}
\caption{Functional sensitivity and layer interactions.
(a) Bypassing StarVLA L6 causes large action deviation despite high
adjacent-layer CKA~\cite{kornblith2019cka}.
(b) Percentage of evaluated layer pairs with joint action deviation below
the sum of individual deviations.
(c) Partial Spearman correlations with action sensitivity. Filled markers
show readout deviation controlling for initial deviation; hollow markers
reverse these roles.
CKA uses calibration inputs; pairwise measurements use
120, 128, and 60 examples for OpenVLA-OFT, StarVLA, and FastWAM, respectively.}
\label{fig:layer_evidence}
\end{figure*}
\subsection{Comparison with Existing Methods}

On OpenVLA-OFT Object, AutoPruner meets the 90\% deployment requirement
with 92.6\% success. At this operating
point, it achieves the lowest measured latency, GFLOPs, and peak GPU memory
in Table~\ref{tab:efficiency_comparison}.
Physical layer removal reduces memory to 4.10 GiB, compared with
14.50--15.95 GiB for the listed runtime methods.
Physical pruning therefore reduces both backbone computation and model
footprint, addressing latency and memory constraints together.
This result illustrates the deployment value of selecting pruning and recovery configurations under an explicit success requirement for resource-constrained robot deployment.

%% file: sections/analysis.tex
\section{Analysis}

\subsection{Functional Evidence for Layer Selection}
\label{sec:alternative-selection}

Figure~\ref{fig:layer_evidence}(a) shows how high representation similarity
can conceal action-sensitive layers. StarVLA L6 has adjacent-layer CKA of
0.9986 yet produces a large action deviation when bypassed. CKA ranks it
first for deletion, which reduces task success below 90\%. FastWAM L0
likewise remains action-sensitive at CKA 0.9995. By measuring action changes
directly, functional probes identify sensitive layers that appear redundant
under CKA.

We next assess layer-selection performance by comparing
functional sensitivity with CKA~\cite{kornblith2019cka}, random orderings,
and rankings generated by language models from architecture
descriptions alone. Each language model produces one ordering without
experimental feedback. Using 200 episodes per configuration and no recovery,
we evaluate progressively longer prefixes and record the largest tested
deletion size retaining at least 90\% task success.
Without recovery, this comparison assesses how well each ordering preserves
task performance directly after deletion.

\begin{figure}[t]
\centering
\includegraphics[width=\columnwidth]{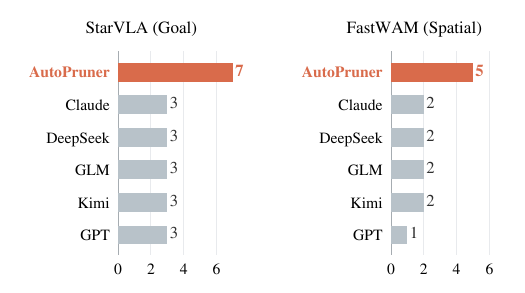}
\caption{Functional selection versus architecture-only language-model rankings
on StarVLA Goal and FastWAM Spatial. Bars show the largest tested deletion
size with at least 90\% success, using 200 episodes per configuration
and no recovery. The highlighted bars use functional sensitivity; each
language model supplies one ordering without experimental feedback.}
\label{fig:agent_backend_selection}
\end{figure}

As shown in Figure~\ref{fig:agent_backend_selection}, functional selection
supports deeper pruning than all five architecture-only rankings on StarVLA
Goal and FastWAM Spatial. An additional OpenVLA-OFT Long comparison yields
four versus at most two deletions. These results show that measured action
effects provide useful selection evidence beyond architecture descriptions.

A fixed-depth comparison complements the threshold-based results by matching the number of deleted layers across selectors for a direct comparison.
At eight deletions, functional selection achieves 72.0\% success on
OpenVLA-OFT Long, compared with 44.0\% for CKA and 0.0\% for random
selection. On FastWAM Object, it achieves 53.0\%, compared with 0.0\%
for CKA and 35.0\% for random selection.
Each deployment uses one random mask. CKA permits one more deletion at the 90\% threshold on
OpenVLA-OFT Long, but functional selection better preserves performance
at the tested eight-layer depth.

\subsection{Layer Interactions and Recovery}

Figure~\ref{fig:layer_evidence}(b) shows that joint action deviation falls
below the sum of individual deviations for 93.8--98.8\% of evaluated pairs.
Thus, summing individual deviation magnitudes overestimates the
joint deviation in these measurements.

Figure~\ref{fig:layer_evidence}(c) helps explain why action-level probing is
informative. Readout deviation remains associated with action sensitivity
after controlling for initial deviation, with partial Spearman correlations
of 0.660--0.855; the reverse associations are much weaker.
The relevant signal is therefore how strongly a perturbation reaches the
action readout, rather than its initial magnitude alone.

\begin{figure}[t]
\centering
\includegraphics[width=\columnwidth]{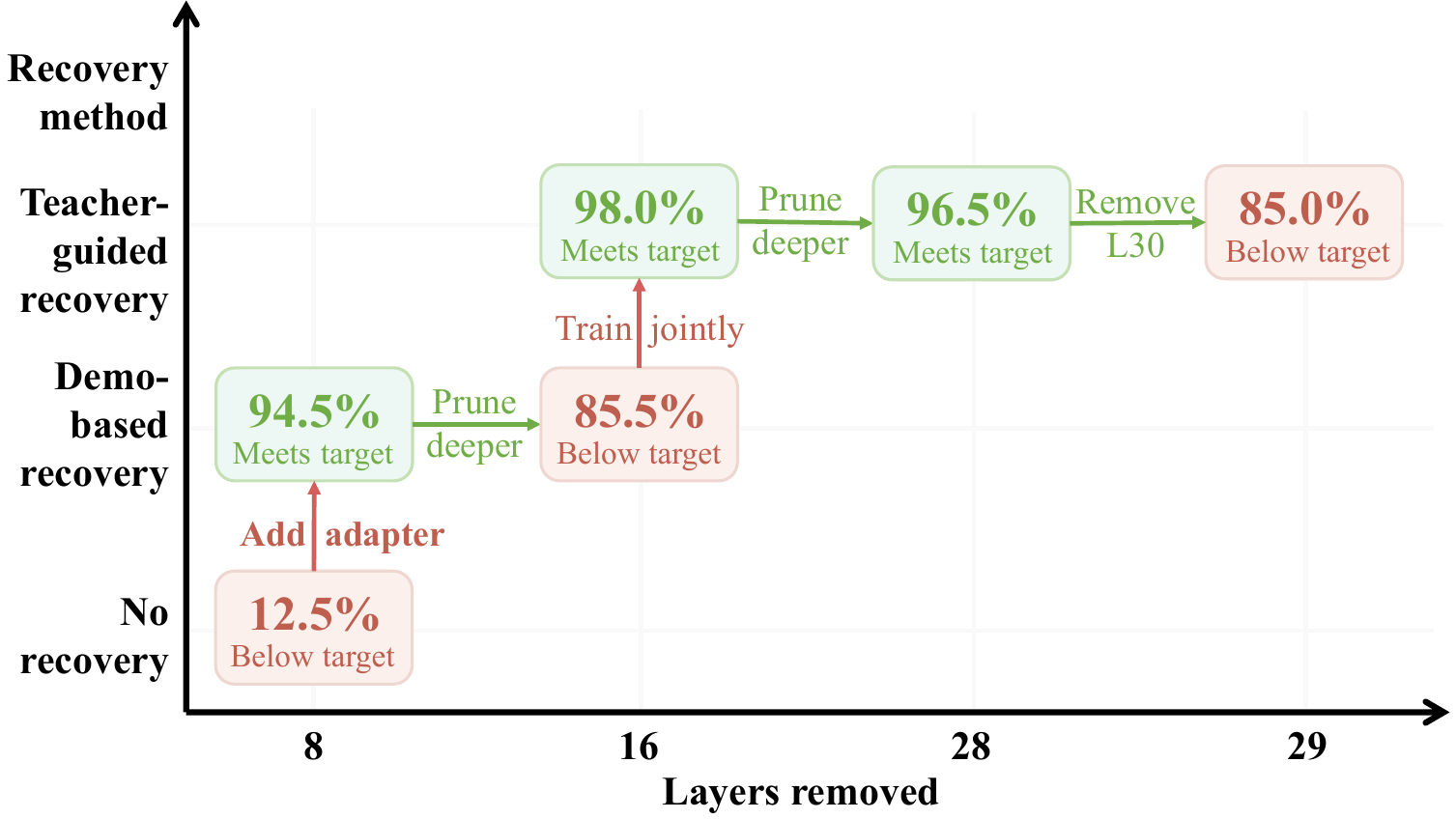}
\caption{Recovery ablations on OpenVLA-OFT Object using 200 OSMesa episodes
per configuration. Vertical arrows vary recovery at fixed deletion sets;
horizontal arrows increase pruning depth. From left to right, the masks remove
L22--L29, L8--L23, L2--L29, and L2--L30. The dashed transition also changes
the adapter recipe.}
\label{fig:recovery_ablation}
\end{figure}

Recovery expands the feasible pruning depth, as shown in
Figure~\ref{fig:recovery_ablation}.
These recovery ablations use 200 episodes per configuration with OSMesa rendering.
On OpenVLA-OFT Object, recovery makes the fixed eight-layer deletion set
feasible, showing that low success before recovery need not disqualify a
structure. At sixteen deleted layers, switching
from adapter-only to joint recovery (adapter, action head, and proprioceptive
projector) raises success from 85.5\% to 98.0\%.
With joint recovery, deleting L2--L29 achieves 96.5\% success, whereas
additionally deleting L30 reduces success to 85.0\%. Thus, the feasibility of a deletion set cannot be judged independently of its recovery recipe.

Recovery also changes the relative performance of deletion sets.
On OpenVLA-OFT Spatial, deleting the eight least action-sensitive layers
achieves 85.8\% success without recovery, compared with 13.8\% for the
AutoPruner configuration removing L4--L11. After both configurations undergo
10000 steps of the same joint recovery recipe, AutoPruner reaches
92.8\% [90.2, 94.8] success, compared with 91.6\% [88.8, 93.7]
for sensitivity-greedy selection. Brackets give 95\% Wilson confidence
intervals in percent. Each result uses 500 episodes with
EGL rendering. The observed reversal shows that higher success before
recovery need not identify the better configuration after recovery.

A separate OpenVLA-OFT Object search illustrates how GPT-5.5 uses
evaluation feedback to revise a proposal. After deleting 27 layers
(L3--L29) yielded 88.5\% success, the agent proposed retaining L29
while keeping the joint recovery recipe fixed. The deletion sets differed
only in L29. The subsequent 26-layer trial (L3--L28) recorded 93.5\%
success, exceeding the 90\% threshold. Both configurations used 10000
recovery steps and 200 evaluation episodes each.
Whereas the recovery ablations vary training at fixed structures, this case
shows the agent using feedback to revise the structure at a fixed recovery recipe.
Recovered representations vary with the recovery recipe. OpenVLA-OFT
policies regain task performance while retaining substantial feature
differences, whereas StarVLA recipes with hidden-state supervision yield
representations closer to the teacher.

%% file: sections/discussion.tex
\section{Discussion}

The experiments indicate that pruning depth and recovery requirements do not transfer uniformly across models or task suites. A configuration that is feasible for one deployment can fail for another, even with the same backbone. This favors deployment-specific search over a single fixed pruning recipe. The resulting compression configuration is specific to a deployment,
while the discovery and verification procedure can be applied across models.

Depth pruning and runtime acceleration address different deployment
constraints. Runtime methods reduce inference computation while typically
retaining all parameters~\cite{xu2025vlacache,zhang2025molevla}, whereas depth pruning physically shortens
the backbone and reduces the model footprint~\cite{men2025shortgpt,yang2025efficientvla}.
Runtime techniques may further accelerate AutoPruner-compressed models,
with their combined effect assessed through closed-loop evaluation.

Our current evaluation focuses primarily on simulated deployments and depth
pruning. Real-robot experiments are needed to assess robustness to sensing
noise, actuation error, and timing variation. Extending AutoPruner to additional
compression dimensions, such as width reduction and token pruning, would enlarge
the search space and introduce new interactions with recovery. Improving the
efficiency of candidate evaluation will therefore be important for scaling the
procedure to broader compression settings.

%% file: sections/conclusion.tex
\section{Conclusion}

We presented AutoPruner, an agent-driven closed-loop framework for
deployment-specific compression of vision-language-action models. Its agent
uses functional layer evidence, search history, and closed-loop task feedback
to adapt pruning structures and recovery strategies to each deployment.
Experiments demonstrate reductions in model size and inference cost,
with deployment-specific trade-offs in task performance. Our analysis shows that
representation similarity can obscure functionally important layers, that
pruning effects depend on downstream interactions, and that recovery can
substantially shift the boundary of feasible compression. These results support
joint search over structure and recovery as a practical approach to VLA
compression, with task feedback guiding deployment configuration.